\documentclass[10pt,twocolumn,letterpaper, table,xcdraw,dvipsnames]{article}

\usepackage{cvpr}              
\usepackage{CJKutf8}
\usepackage{multirow}
\usepackage[ruled,linesnumbered]{algorithm2e}

\usepackage{graphicx}
\usepackage{epstopdf}
\usepackage{float}

\usepackage[dvipsnames]{xcolor}

\definecolor{cvprblue}{rgb}{0.21,0.49,0.74}
\usepackage[pagebackref,breaklinks,colorlinks,citecolor=cvprblue]{hyperref}

\def\paperID{8} 
\def\confName{CVPR}
\def\confYear{2024}

\title{Attack End-to-End Autonomous Driving through Module-Wise Noise}

\author{Lu Wang$^{1,3}$, Tianyuan Zhang$^{1,2,3}$, Yikai Han$^{1}$, Muyang Fang$^{1}$, Ting Jin$^{1}$, Jiaqi Kang$^{4}$\\
$^1$ School of Computer Science and Engineering, Beihang University, Beijing, China\\
$^2$ Shen Yuan Honors College, Beihang University, Beijing, China \\
$^3$ State Key Lab of Software Development Environment, Beihang University, Beijing, China\\
$^4$ School of Software, Beihang University, Beijing, China\\
{\tt\small $\{$20373361, zhangtianyuan, 21371441, 21373061, 21371466, 21373331$\}$@buaa.edu.cn}}

\begin{document}
\begin{CJK}{UTF8}{gbsn}
\maketitle
\begin{abstract}
With recent breakthroughs in deep neural networks, numerous tasks within autonomous driving have exhibited remarkable performance. However, deep learning models are susceptible to adversarial attacks, presenting significant security risks to autonomous driving systems. Presently, end-to-end architectures have emerged as the predominant solution for autonomous driving, owing to their collaborative nature across different tasks. Yet, the implications of adversarial attacks on such models remain relatively unexplored. In this paper, we conduct comprehensive adversarial security research on the modular end-to-end autonomous driving model for the first time. We thoroughly consider the potential vulnerabilities in the model inference process and design a universal attack scheme through module-wise noise injection. We conduct large-scale experiments on the full-stack autonomous driving model and demonstrate that our attack method outperforms previous attack methods. 
We trust that our research will offer fresh insights into ensuring the safety and reliability of autonomous driving systems.
\vspace{-0.20in} 
\end{abstract}    
\section{Introduction}
\label{sec:intro}

\begin{figure}
    \centering
    \vspace{-0.15in}   \includegraphics[width=0.96\linewidth]{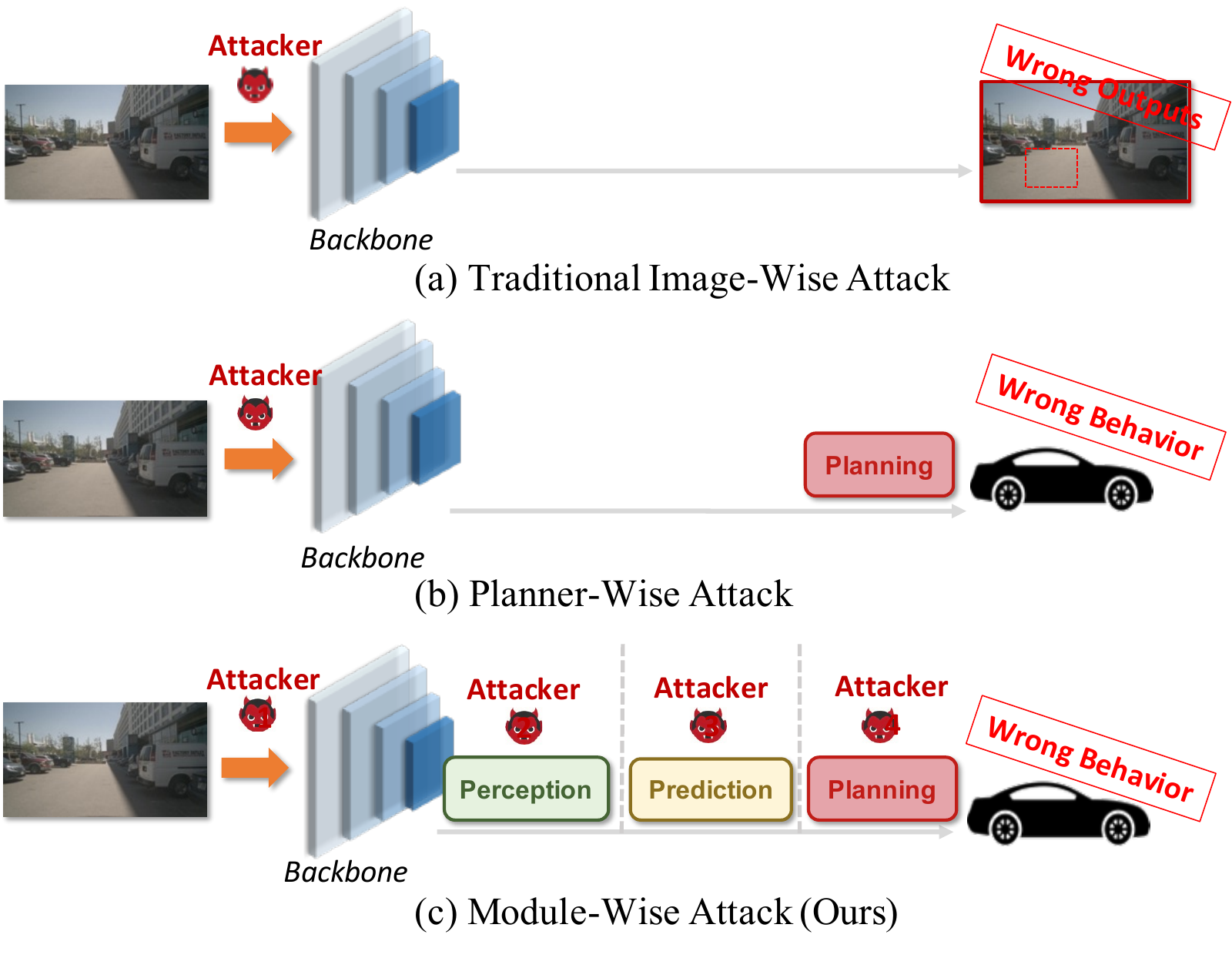}
    \caption{Adversarial attacks in autonomous driving. There are a considerable number of mature attack algorithms targeting the perception of autonomous driving (\textbf{a}). There is a limited amount of research focusing on adversarial security for end-to-end regression-based decision models (\textbf{b}). We propose the module-wise attack targeting end-to-end autonomous driving models (\textbf{c}).}
    \vspace{-0.20in} 
    \label{fig:head}
\end{figure}

With recent significant advancements in deep learning, autonomous driving technology plays an increasingly important role in today's society.
End-to-end autonomous driving models map raw sensor data directly to driving decisions, becoming the predominant solution gradually. However, despite the excellent performance of end-to-end models, we must also realize the security challenges they face. A considerable amount of research has already been devoted to studying adversarial attack methods towards individual tasks within the field of autonomous driving, with particular emphasis on the perception layer \cite{abdelfattah2021towards, cao2021invisible}, as illustrated in Figure \ref{fig:head} (a). Nowadays, researchers have begun to conduct adversarial attacks on very simple end-to-end regression-based decision models \cite{wu2023adversarial} directly from input images, 
as illustrated in Figure \ref{fig:head} (b).

However, there is currently no adversarial security research conducted on complex end-to-end autonomous driving models that are composed of multiple sub tasks. 
In this paper, we delve into the robustness of modular end-to-end autonomous driving models. 
We believe that attacking complex models should not only focus on the image level but also consider the vulnerability of the interaction process between modules, as illustrated in Figure \ref{fig:head} (c). Therefore, we introduce adversarial noise at the interfaces between modules of the end-to-end models. 
The contributions of this paper are summarized as follows:

\begin{itemize}
    
    \item We design the module-wise attack toward end-to-end autonomous driving models.
    
    \item We conduct extensive experiments on the full-stack autonomous driving model 
    , which reveals the insecurity of the model.
  
\end{itemize}

\section{Related Work}
\label{sec:formatting}

\subsection{End-to-End Autonomous Driving}


The early end-to-end autonomous driving models combine relatively few tasks. \cite{zeng2019end} adopts a bounding box-based intermediate representation to construct the motion planner. Considering that Non-Maximum Suppression(NMS) in this method can lead to loss of perceptual information, P3 \cite{p3} innovatively designs an end-to-end network that utilizes maps, advanced control instructions, and LIDAR points to generate interpretable intermediate semantic occupancy representations, which facilitates safer trajectory planning. Following P3, MP3 \cite{mp3} and ST-P3 \cite{stp3} achieve new improvements. 
UniAD \cite{uniad} is the first end-to-end network that integrates full-stack autonomous driving tasks. By thoroughly considering the contributions of each task to autonomous driving and mutual promotion among modules, UniAD significantly surpasses previous sota performance on each task.

\subsection{Adversarial Attack}

\begin{figure*}[!t]
    \centering
    \vspace{-0.15in}
    \includegraphics[width=0.98\linewidth]{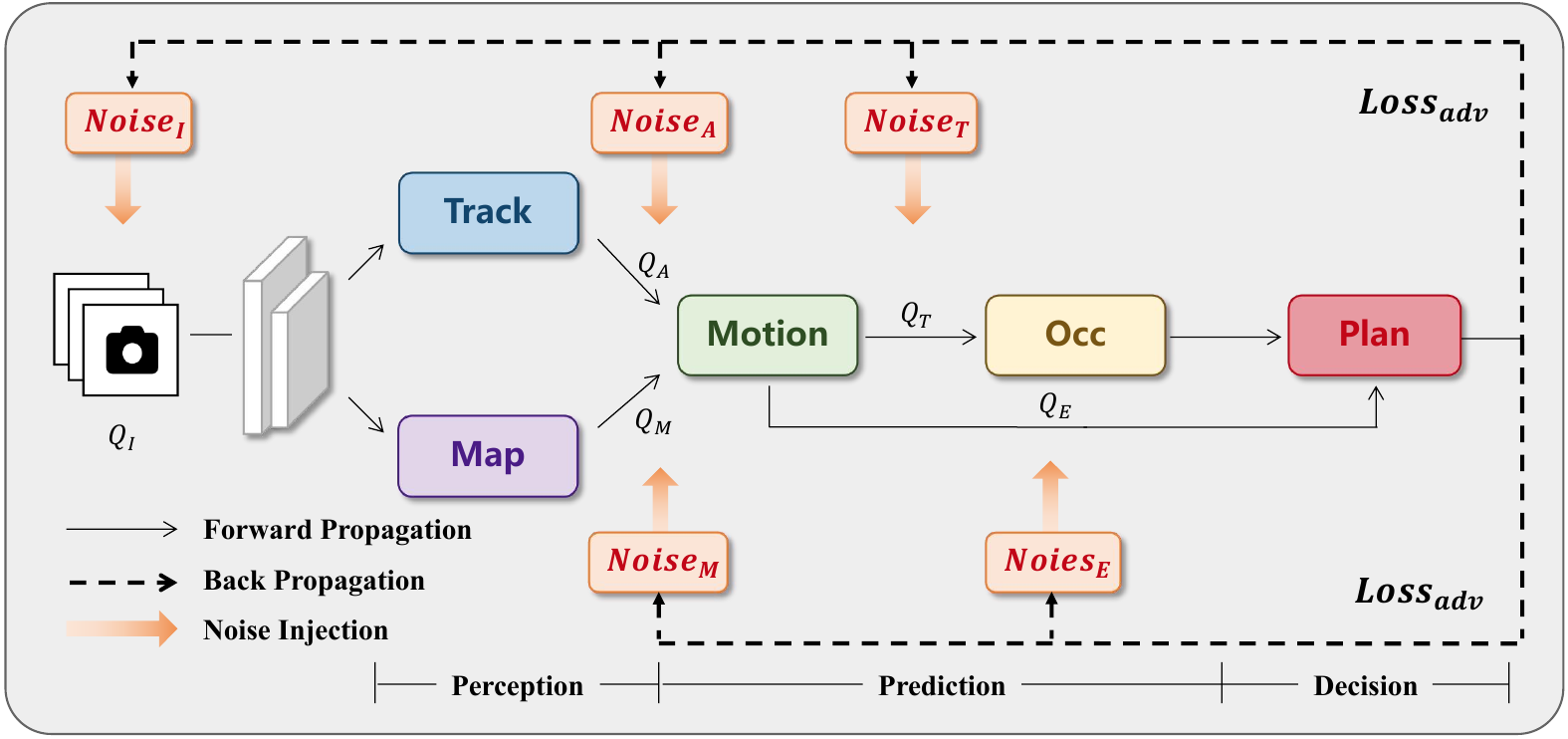}
    \caption{The framework of our module-wise noise attack method. We meticulously inject adversarial noise into the interaction process of all modules in the end-to-end autonomous driving model and synchronize the optimization of all noise using the losses from all modules.}
    \label{fig:framework}
    \vspace{-0.15in}
\end{figure*}

Adversarial noise refers to carefully crafted perturbations designed for neural network input data, which are typically small but can cause models to produce complete error outputs \cite{szegedy2013intriguing,liu2019perceptual,liu2020bias,liu2023x,liu2020spatiotemporal,wang2021dual,liu2022harnessing,liu2023exploring,tang2021robustart}. \cite{szegedy2013intriguing} first introduces the concept of adversarial attack and utilizes L-BFGS approximation. 
Following that, a series of adversarial attack methods are proposed, such as gradient-based methods \cite{dong2018boosting, madry2017towards}, and optimization-based methods \cite{carlini2017towards}. Although early adversarial attacks primarily target image classification models, they demonstrate the vulnerability of neural networks. Their attack principles have guided the implementation of various attack methods in different tasks, posing a serious threat to the practical application of models in the real world\cite{wang2021dual, liu2019perceptual, liu2020bias}.

\section{Methodology}


\subsection{Problem Definition}
\label{sec:pd}

Let $y = F(x; \theta)$ represent the end-to-end autonomous driving model with parameters $\theta$, which includes $n$ sub modules $\{M_{i}\}_{i=1}^{n}$. The inference stage can be formulated as:
\begin{equation}
    Q_{i} = M_{i}( Q_{i-1}; \theta_{i} ), i = 1, 2, ..., n,
\end{equation}
where $Q_{0} = x$ refers to the input images, $Q_{n} = y$ the planned results, and $Q_{i}$ the output intermediate queries of the $i$-th sub module.

In this research, we consider the interaction information among modules as a potential vulnerability. In addition to the input images, we inject adversarial noise $N_{i-1}$ into the input data $Q_{i-1}$ of each submodule $M_{i}$. That is, 
\begin{equation}
    Q_{i-1}^{adv} = Q_{i-1} + N_{i-1}, i = 1, 2, ..., n.
\end{equation}

Let $N$ represents the final injected adversarial noise $\{N_{i}\}_{i=0}^{n-1}$, and $y^{adv} = F(x; \theta, N)$ denotes the decision results made by the end-to-end model after inference through each module under noise attack. Our attack objective is to find, for each data set $x$, an adversarial noise $N$ that satisfies the constraint on the perturbation magnitude $\xi$, while ensuring $y^{adv} \ne y$. 

\subsection{Module-Wise Adversarial Noise}
\label{sec:mwnoise}

Figure \ref{fig:framework} presents an overall framework of the proposed method. A complete autonomous driving model comprises perception, prediction and the final planner. We aim to inject adversarial noise into the input of all sub modules in the end-to-end autonomous driving model so mainly focus on the current full-stack autonomous driving model UniAD \cite{uniad}. Undoubtedly, our noise injection scheme is applicable to any modular end-to-end autonomous driving model. 

Images serve as the initial input for the model to perceive the environment, forming the foundation for all subsequent modules. Therefore, we design pixel-level noise specifically for input images, denoted as $N_{I}$. The track module accomplishes multi-object tracking, producing spatiotemporal information $Q_{A}$. Therefore, we design noise $N_{A}$ for the spatiotemporal features of agents. The mapping module models the features of road elements as $Q_{M}$ and we design adversarial noise $N_{M}$ for all map features. $Q_{A}$ and $Q_{M}$ respectively provide dynamic agent features and static scene information for downstream modules. 
The motion module, based on the aforementioned dynamic and static features, predicts the most likely future trajectory states of all agents, represented as $Q_{T}$. Similarly, we design noise $N_{T}$ for the state of agents. 
In addition, the motion module also expresses the future intentions of the ego vehicle, represented as $Q_{E}$. We separately design noise $N_{E}$ for $Q_{E}$. Since the occupancy map is ultimately used only in post-processing for collision optimization of the planned trajectory, and this optimization process is non-differentiable, we don't consider injecting noise into the occupancy map.


\subsection{Attack Strategy}
\label{sec:attack}

    
        





        

We achieve the attack by iteratively optimizing module-wise noise. To be more specific, during the model's inference stage for each batch of data, we first initialize corresponding adversarial noise randomly within the perturbation constraints for each module, following the forward propagation process of the model. At each iteration, noise injected into the corresponding module is propagated and stored until the final planning stage. We synchronize the update of all noise by the adversarial loss, using it to initialize the noise for the next iteration.

Adversarial loss, essential for noise update, includes attack loss and noise loss. Attack loss seeks to maximize deviation between sub module predictions and actual results, mathematically represented as:
\begin{equation}
    L_{att} = L_{track} + L_{map} + L_{motion} + L_{occ} + L_{plan}.
\end{equation} 
The noise loss is used to constrain the injected adversarial loss to be as minimal as possible. We choose the L2 norm distance to measure the magnitude of module-wise noise. It is represented as:
\begin{equation}
    L_{noi} = \sum_{i} \sigma_{i} \cdot ||N_{i}||_{2}, i = I, A, M, T, E,
\end{equation} 
where $\sigma_{i}$ are hyper-parameters. The final adversarial loss is represented as:
\begin{equation}
    \label{eq:loss}
    L_{adv} = L_{att} - L_{noi}.
\end{equation}

We optimize the noise in a manner similar to PGD
, performing gradient ascent on the adversarial noise, namely:
\begin{equation}
    \label{eq:update}
    N_{i}^{t+1} = \prod_{\epsilon} (N_{i}^{t} + \frac{\epsilon}{\sqrt{k}} \cdot sgn(\nabla_{Ni} L_{adv})),
\end{equation} 
where k denotes the number of iterations and $\epsilon$ is a hyperparameter that constrains the magnitude of the perturbation. 

\section{Experiments}

\subsection{Experimental Setup}

\quad \textbf{Dataset.} Our robustness evaluation experiments are conducted on the validation split of the large-scale autonomous driving dataset nuScenes \cite{caesar2020nuscenes}. 

\textbf{Model.} For the target model, we choose the current full-stack autonomous driving model UniAD, which includes complete sub tasks of perception, prediction, and decision. We conduct noise injection according to the five stages of the model as outlined in Section \ref{sec:mwnoise}. For the specific implementation of the attack, we set the number of iterations $k = 10$, $\sigma_{I} = 8 \times 10^{-6}$, $\sigma_{A} = \sigma_{M} = \sigma_{T} = \sigma_{E} = 2 \times 10^{-4}$. 

\textbf{Metrics.} For the metrics, we choose the evaluation methods for five sub tasks adopted in \cite{uniad}.

\begin{table*}[t]
\centering
\renewcommand\arraystretch{1.5}
\caption{Comparison with other attack methods on five modules. The first row represents the original replication results. The map module is evaluated by IOU of road elements. The motion module is evaluated by three types of error on the vehicle. The plan module is evaluated by the average of L2 error and collision rate in the next three seconds.  Our attack method
(the last row) achieves the maximum performance degradation across all modules.}
\label{tab:compa-five}
\huge
\resizebox{\textwidth}{!}{%
\begin{tabular}{@{}c|cccc|ccc|ccc|cc|cc@{}}
\toprule[2.5pt]
                         & \multicolumn{4}{c|}{Track}                                                                           & \multicolumn{3}{c|}{Map}                                                             & \multicolumn{3}{c|}{Motion}                                                         & \multicolumn{2}{c|}{Occupancy}                              & \multicolumn{2}{c}{Plan}                                                 \\
\multirow{-2}{*}{Method} & \cellcolor[HTML]{EFEFEF}Amota $\uparrow$ & Amotp $\downarrow$ & Recall $\uparrow$ & IDS $\downarrow$ & Drivable $\uparrow$ & \cellcolor[HTML]{EFEFEF}Lanes $\uparrow$ & Crossing $\uparrow$ & \cellcolor[HTML]{EFEFEF}minADE $\downarrow$ & minFDE $\downarrow$ & MR $\downarrow$ & \cellcolor[HTML]{EFEFEF}Iou-n $\uparrow$ & Iou-f $\uparrow$ & \cellcolor[HTML]{EFEFEF}L2 error(m) $\downarrow$ & Col.Rate $\downarrow$ \\ \midrule
Original                 & \cellcolor[HTML]{EFEFEF}0.367            & 1.26               & 0.449             & 442              & 68.80\%             & \cellcolor[HTML]{EFEFEF}31.17\%          & 14.29\%             & \cellcolor[HTML]{EFEFEF}0.733               & 1.079               & 0.164           & \cellcolor[HTML]{EFEFEF}64.00\%          & 41.00\%          & \cellcolor[HTML]{EFEFEF}1.09                     & 0.32\%                \\ \midrule
Image-specific Attack    & \cellcolor[HTML]{EFEFEF}0.008            & 1.947              & \textbf{0.051}    & 325              & 45.25\%             & \cellcolor[HTML]{EFEFEF}17.34\%          & 4.42\%              & \cellcolor[HTML]{EFEFEF}1.424               & 2.110               & 0.265           & \cellcolor[HTML]{EFEFEF}24.90\%          & 12.60\%          & \cellcolor[HTML]{EFEFEF}2.90                     & 2.27\%                \\
Image-agnostic Attack    & \cellcolor[HTML]{EFEFEF}0.231            & 1.901              & 0.095             & 224              & 49.67\%             & \cellcolor[HTML]{EFEFEF}19.96\%          & 6.22\%              & \cellcolor[HTML]{EFEFEF}1.119               & 1.650               & 0.226           & \cellcolor[HTML]{EFEFEF}32.90\%          & 17.20\%          & \cellcolor[HTML]{EFEFEF}1.31                     & 0.45\%                \\ \midrule
Module-wise Attack       & \cellcolor[HTML]{EFEFEF}\textbf{0.000}   & \textbf{1.976}     & \textbf{0.051}    & \textbf{2706}    & \textbf{39.63\%}    & \cellcolor[HTML]{EFEFEF}\textbf{15.27\%} & \textbf{2.89\%}     & \cellcolor[HTML]{EFEFEF}\textbf{2.985}      & \textbf{4.914}      & \textbf{0.409}  & \cellcolor[HTML]{EFEFEF}\textbf{17.60\%} & \textbf{8.00\%}  & \cellcolor[HTML]{EFEFEF}\textbf{5.37}            & \textbf{4.33\%}       \\ \bottomrule[2.5pt]
\end{tabular}%
}
\vspace{-0.15in}
\end{table*}

\subsection{Robust Evaluation}

\quad \textbf{Baselines.} As there are currently no adversarial attacks targeting complex end-to-end autonomous driving models composed of a series of sub modules, we choose attack methods targeting end-to-end regression-based decision models as baselines \cite{wu2023adversarial}. 
In our comparative experiments, we set the maximum allowable perturbation $\epsilon$ for images to 8 and implement \textit{Image-specific Attack} and \textit{Image-agnostic Attack} tailored for UniAD.

\textbf{Results.} 
We provide the robustness evaluation results compared with baselines for the five tasks, as shown in Table \ref{tab:compa-five}.  Our attack can significantly reduce the performance of all tasks. We measure the main planning performance of the vehicle using the L2 error between the planned trajectory and the ground truth trajectory, as well as the collision rate with obstacles in the ego vehicle's driving environment. Results indicate that all three attacks lead to errors in the vehicle's planning, but the proposed attack method poses the greatest threat to the model as it interferes with both the perception and prediction processes during model inference, and the final prediction heavily relies on the inference results of upstream modules, resulting in severe planning errors due to error accumulation. Overall, the attack intensities of the three methods exhibit the same trend across the five tasks, with the \textit{Image-specific Attack} being much stronger than the \textit{Image-agnostic Attack}, and our module-wise attack surpassing the \textit{Image-specific Attack}.


\section{Conclusions}

We delve into the robustness of complex end-to-end autonomous driving models with multi-modules by introducing a novel adversarial attack using module-wise noise injection. We strongly believe that it is imperative to consider the vulnerabilities in the interaction process between modules to enhance the security of autonomous driving systems. By conducting extensive experiments on the full-stack autonomous driving model, we demonstrate the profound impact of injecting noise into different modules on the planning performance of the model as well as other tasks. 


\section{Acknowledgment}

This work is supported by grant No. KZ46009501.
{
    \small
    \bibliographystyle{ieeetr} 
    \bibliography{main}
}

\end{CJK}
\end{document}